\documentclass[10pt]{article} % For LaTeX2e
\usepackage[preprint]{tmlr}
\usepackage{amsmath,amsfonts,bm}

\def\eqref#1{equation~\ref{#1}}
\def\1{\bm{1}}

\DeclareMathAlphabet{\mathsfit}{\encodingdefault}{\sfdefault}{m}{sl}
\SetMathAlphabet{\mathsfit}{bold}{\encodingdefault}{\sfdefault}{bx}{n}

\usepackage{hyperref}
\usepackage{url}
\usepackage{amsmath}
\usepackage{booktabs}
\usepackage{multirow}
\usepackage{float}
\usepackage{graphicx}
\usepackage{placeins}

\title{SortBench: Benchmarking LLMs based on their ability to sort lists}

\author{\name Steffen Herbold \email steffen.herbold@uni-passau.de \\
       \addr Faculty of Computer Science and Mathematics\\
       University of Passau\\
       Passau, Germany}

\def\month{MM}  % Insert correct month for camera-ready version
\def\year{YYYY} % Insert correct year for camera-ready version
\def\openreview{\url{https://openreview.net/forum?id=XXXX}} % Insert correct link to OpenReview for camera-ready version

\begin{document}

\maketitle

\begin{abstract}
Sorting is a tedious but simple task for human intelligence and can be solved fairly easily algorithmically. However, for Large Language Models (LLMs) this task is surprisingly hard, as some properties of sorting are among known weaknesses of LLMs: being faithful to the input data, logical comparisons between values, and strictly differentiating between syntax (used for sorting) and semantics (typically learned by embeddings). Within this paper, we describe the new SortBench benchmark for LLMs that comes with different difficulties and that can be easily scaled in terms of difficulty. We apply this benchmark to seven state-of-the-art LLMs, including current test-time reasoning models. Our results show that while the o3-mini model is very capable at sorting in general, even this can be fooled if strings are defined to mix syntactical and semantical aspects, e.g., by asking to sort numbers written-out as word. Furthermore, all models have problems with the faithfulness to the input of long lists, i.e., they drop items and add new ones. Our results also show that test-time reasoning has a tendency to overthink problems which leads to performance degradation. Finally, models without test-time reasoning like GPT-4o are not much worse than reasoning models. 
\end{abstract}

\section{Introduction}

Sorting is a basic skill that humans acquire at an early age already: we learn the order of numbers, the alphabet, and we also learn to sort by shapes or similar aspects. While sorting with very large numbers of items is a tedious activity for humans, we can solve it reliably nonetheless. Moreover, algorithmic sorting strategies are well-known and range from intuitive (bubble sort, insertion sort) to more optimized variants (quicksort, mergesort). Thus, from both a human intelligence perspective as well as from the algorithmic complexity required to solve the problem, sorting is a simple problem. 

Modern Large Language Models (LLMs) based on decoder-only transformers~\citep{radford2018improving}, instruction fine-tuning~\citep{zhang2023instruction}, and even advanced chain-of-thought reasoning~\citep{jaech2024openai} are showing increasingly strong capabilities at even solving hard problems like advanced math~\citep{artofproblemsolvingProblemSolving} or the ARC challenge~\citep{arcprizePrize}. Still, we believe that the simple task of sorting a list is still hard and yet unsolved for such models. Concretely, we believe that sorting has several properties that make this both a hard challenge for LLMs and a good benchmark to understand important LLM qualities like faithfulness to the input, logical comparisons, and the capability to strictly differentiate between the syntax of the input (relevant for sorting) and its semantics (typically learned by embeddings). 

The idea of SortBench is simple: use a randomized generator to create lists, give these lists to an LLM with a prompt that instructs it to sort them, and then evaluate if the lists were sorted correctly and returned in the expected output format. The evaluation considers whether the order of list items is correct and if the items in the returned list are the same as in the unsorted list. Using different generators for the lists allows us to evaluate different properties, e.g., sorting of integers, floats, random strings, and words for lists of different lengths. The difficulty can be further increased by creating variants of the above, e.g., requiring an lexicographic sorting of words that represent numbers (e.g., one, two, three). The benchmark is designed to assess both the general sorting capabilities of LLMs and also to understand which properties of data make sorting more difficult for them, whether they follow the instructions of the prompt, and whether they manage to leave the content of the list unchanged. 

\section{Method}

Below, we describe the generation of the SortBench v1.0 benchmark for the basic and advanced difficulty, as well as debug tasks that are useful to better understand the LLM's capabilities. SortBench focuses on lists over two input domains, i.e., sorting numbers written in decimal notation and lexicographic sorting of strings. For the strings, we do not consider special characters, but rather restrict the benchmark to ASCII letters and non-English words. Further, we write all lists in the format as Python lists, i.e., with opening and closing brackets \texttt{[]}, numbers without quotation marks and strings within quotation marks.

\subsection{Prompt}

Our benchmark is designed to evaluate the zero-shot capabilities of LLMs without any fine-tuning neither through modifying weights nor through prompt engineering. While specifically crafted and engineered prompts, including few-shot prompts, may lead to better results, a basic capability like sorting should not require such considerations. Hence, we only give the models a simple, straightforward prompt telling them what to do, without any details on the exact, lengthy specifications or data formats, or similar. We use the following system and user prompts:
\begin{itemize}
    \item System prompt: ``Your task is to sort a list according to the common sorting of the used data type in Python. The output must only contain the sorted list and nothing else. The format of the list must stay the same.''
    \item User prompt: ``Sort the following list: <list>'' where <list> is replaced with the list that needs to be sorted in Python, e.g., \texttt{[16, 5, 10]} for an integer list or \texttt{['foo', 'bar', 'baz']} for a list of strings. 
\end{itemize}

\subsection{Benchmark tasks}

The tasks within our benchmark are grouped by their difficulty. We have basic tasks, that should align well with how LLMs are trained, advanced tasks that rather push the LLMs to cases that might be problematic, and debug tasks that can be used for additional diagnostics with respect to how well LLMs follow the general instruction of sorting a list without changes to the list's items. 

\subsubsection{Basic difficulty}

For the basic difficulty, we generate lists that are simple in the sense that their contents should align fairly well with how LLMs are trained, specifically, list items they likely see often during the training. This is based on the assumption that logical comparisons are easier for LLMs between tokens that were seen during training. Further, we avoid aspects that might confuse the sorting, e.g., having the same list item multiple times or using negative numbers. Thus, the basic difficulty evaluates the general capability to follow the instruction to sort a list, since there are no possible confusing constructs in the tasks and all list items are from the same input domain (i.e., numbers or words). Based on these goals, we use the following three types of lists for the basic difficulty: 
\begin{itemize}
    \item \textbf{Int-0:1000}: Integers in the range 0 to 1,000 without duplicates. 
    \item \textbf{Float-0:1000}: 32-bit floating point numbers printed a full precision decimal notation in the range 0 to 1,000 without duplicates. 
    \item \textbf{English}: Words from the English language sampled from the WordNet corpus~\citep{fellbaum2010wordnet} without duplicates. Further, we disallow words with an apostrophe to avoid potential issues with with Python's string parsing. 
\end{itemize}

\subsubsection{Advanced difficulty}

With the advanced difficulty we want to understand if orders are not only memorized for token combinations that may have been known from training, but also to other, more complex, scenarios. Note that we do not consider duplication of list items as part of this difficulty, as we rather consider this as a debug task (see Section~\ref{sec:debug-tasks}). For numbers, we make the task more difficult by looking at uncommonly large values and values that are very close to each other, i.e., compressed within a small interval. Further, we add lists with negative values to measure whether signs are interpreted correctly by the LLMs. This results in the following advanced tasks for numbers: 
\begin{itemize}
    \item \textbf{Int-10000000:10001000}: Integers in the range 10,000,000 to 10,001,000 without duplicates.
    \item \textbf{Float-10000000:10001000}: 32-bit floating point numbers printed at full precision in decimal notation in the range 10,000,000 to 10,001,000 without duplicates.
    \item \textbf{Float-0:0.0001}: 32-bit floating point numbers printed at full precision in decimal notation in the range 0 to 0.0001 without duplicates.
    %\item \textbf{Int-n1000:-1}: Integers in the range -10,000 to -1 without duplicates.
    \item \textbf{Int-n1000:1000}: Integers in the range -1,000 to 1,000 without duplicates.
    %\item \textbf{Float-n1000:0}: 32-bit floating point numbers in the range -10,000 to -1 without duplicates.
    \item \textbf{Float-n1000:1000}: 32-bit floating point numbers in the range -1,000 to 1,000 without duplicates.
    %\item \textbf{Float-n0.0001:0.0001}: 32-bit floating point numbers in the range -0.0001 to 0.0001 without duplicates. 
\end{itemize}

For strings, the easiest way to increase the difficulty is to not use actual words, but rather random strings. However, we can also try to actively confuse the model. For this, we consider two scenarios. First, we add a prefix of equal characters (e.g., ``rrrHello'''). This requires the sorting to basically ignore the beginning of the tokens associated with a list item and sort based on later characters in the string. Second, we try to confuse the LLM by using number words (e.g., ``one'', ``three-thousand''). Through this, we try to trick the LLM as it may rather consider the semantic meaning of the number words and sort by the numbers instead of the lexicographic order. This results in the following advanced tasks for strings: 
\begin{itemize}
    \item \textbf{ascii}: Strings of five random, lowercase ASCII letters without duplicates
    \item \textbf{AsCiI}: Strings of five random ASCII letters with both lower and uppercase letters without duplicates.
    \item \textbf{PrfxEnglish}: Words from the English language without duplicates that have a constant letter repeated three times as prefix. The prefix is the same for all words in a list. 
    \item \textbf{NumberWords}: Words that represent integers between one and 1,000 without duplicates.
\end{itemize}

\subsubsection{Debug tasks}
\label{sec:debug-tasks}

In addition to the tasks above used for scoring LLMs, we also include two types of debug tasks that support the diagnostics of LLMs with respect to their faithfulness to the inputs. Through these, we want to enable further insights into the benchmark scores by providing comparisons to scenarios which require the LLMs to respect properties of the input, independent of sorting. For this, we provide two additional variants for all three tasks from the basic difficulty, i.e., six tasks. In the first variant, the lists are already sorted, which reduces the task of the LLM to just repeat a list. This helps us to see the if LLMs are able to not change the lists in isolation, i.e., regardless of mistakes that may be induced here due to sorting. In the second variant, each list item is duplicated, i.e., appears twice in the list. This requires the LLM to not just recognize the word order, but also repeat words the required number of times. This helps us to understand if the LLMs treat list items as separate entities. Bad performance here indicates that syntactic (and possibly also semantic) similarity between list items may affect the task. 

\subsection{List lengths}

We create for each of the above tasks of eight lists different lengths using an exponential growth with a base of two, such that we have lists with $2^1=2, ..., 2^8=256$ items. For each length, we create 10 lists to account for possible random effects. Since we have three basic tasks, nine advanced tasks, and six debug tasks, this means we have to sort $8\cdot 10\cdot(3+9+6)=1440$ lists. The maximum list length was selected such that typical tokenizers based on Byte Pair Encoding \citep[BPE,][]{sennrich2015neural} or similar have less than or equal to 4096 tokens, incl. the system and user prompt. This ensures that the sorting could happen in a context window of 8192 tokens. 

\subsection{Scores for LLM-Sorting}

When we consider the quality of sorting with LLMs, there are three general aspects to consider, i.e., whether the output is a valid Python list, whether the list items match the original list, and whether the result is correctly sorted. 

\subsubsection{Output validity}

The first challenge when working with LLMs is to ensure that their output follows an expected specification. In our case, this means that the output list is still a valid Python list. To check this, we use Python's \texttt{eval} function that takes as input a string and interprets this as a Python command. With a valid list, this would be directly converted. If that fails, we check if the output generation was possibly prematurely aborted, i.e., the LLM yielded an end-of-sequence token when the list was not yet finished. For that, we first check if the output finishes with a closing bracket. If that is not the case, we manually check the list and look for alternative ways to parse the list. Section~\ref{sec:parsing-errors} presents a detailed list of error cases we found, including their frequency. After parsing, we additionally check if the list contains an ellipsis (...), because this is a special data type in Python, cast all list items to the expected type (e.g., `100' to int), and check if the output was parsable, though not a list, but rather a tuple. Based on these outcomes, we define the $ValidityScore$ as follows:
\begin{itemize}
\item $ValidityScore=1$, if the output is a valid Python list, without ellipsis and of the correct type. 
\item $ValidityScore=0.75$ if the output is valid Python, but contains an ellipsis; the type of the list items needed the be cast; the output was a tuple instead of a list; or if the output was only missing the closing bracket.
\item $ValidityScore=0.5$, if the output contained a list, but this required special parsing (see Section~\ref{sec:parsing-errors}).
\item $ValidityScore=0$ if the output did not contain a parsable list. 
\end{itemize}

\subsubsection{Sorting correctness}

We use two metrics for the correctness of the sorting that would be zero for correctly sorted lists:
\begin{itemize}
    \item $UP$ is the percentage of unordered pairs. We conduct pair-wise comparisons between all list items and compute the ratio of such pairs that are not in order.
    \item $UN$ is the percentage of unordered neighbors. For this, we compute the ratio of directly adjacent list items that are not in order. 
\end{itemize}

With $UP$, we get a global perspective that penalizes the sorting if elements are far off from their current position. Such items will be part of many unordered pairs, increasing the count. With $UN$, we get a local perspective and rather consider if most items have a suitable neighbor, such that a single item in a wrong position has a low impact. We use the mean of these two metrics to define 
\begin{equation}
    Sorting Score=1-\frac{UP+UN}{2}
\end{equation}
for a list.

\subsubsection{Faithfulness}

All metrics considered so far are solely computed based on the output of the LLM, i.e., the result of the sorting, without taking the list that was supposed to be sorted into consideration. However, there are no guarantees that the returned lists actually have the same items as the input: LLMs are known to hallucinate, which in this case means altering the list by either dropping or adding items. To account for this, we define two additional metrics: 
\begin{itemize}
    \item $I^+$ is the ratio of added list items, i.e., the number of items that were added in relation to the length of the original list. 
    \item $I^-$ is the ratio of the missing items, i.e., the number of items that were present in input but not in the original list in relation to the length of the original lists. 
\end{itemize}
Please note that both metrics are computed accounting for duplicates, i.e., if an item appears twice in the input, it also needs to appear twice in the output. Further, we clip $I^+$ at one to avoid possible cases in which a LLM might add more new items, than were originally in the list. Same as above, we use the mean of both metrics to define
\begin{equation}
    FaithfulnessScore=1-\frac{I^++I^-}{2}.
\end{equation}

\subsubsection{Total score}
\label{sec:total-score}

The total score for a list is defined using the three criteria validity, sorting correctness, and faithfulness as
\begin{equation}
    SortBenchScore=ValidityScore\cdot \frac{SortingScore+FaithfulnessScore}{2}.
\end{equation}
Hence, we compute the mean of the sorting and faithfulness scores and penalize this, if the list cannot be parsed. 

All scores so far were computed per list, i.e., for each list in every task. While these scores allow good insights into different strengths and weaknesses of the LLMs, benchmarks should ideally have a single value per model per task as final result for ranking. We compute this using a length-weighted mean over all scores for a task:
\begin{equation}
    ModelScore=\frac{\sum Length\cdot mean(SortBenchScore_{Length})}{\sum Length}.
\end{equation}
The weighting by length ensures that all sequence lengths in the benchmark are considered, while preventing very short, easier-to-sort sequences from dominating the total score. This is important because of the exponential growth of lengths, which means we have more lists with short length than with longer lengths. An alternative interpretation for this definition is that it is the area under the scoring curve with the sequence length on the x-axis and the score on the y-axis. These scores can then be averaged to get the performance per difficulty, as well as overall. 

\subsection{Model selection}
\label{sec:models}

We used SortBench to evaluate a broad range of LLMs. We include current state-of-the art models without test-time reasoning, i.e., GPT-4o and Claude Sonnet 3.5, as well as their smaller counterparts GPT-4o-mini and Claude 3.5 Haiku~\citep{hurst2024gpt, anthropicIntroducingClaude}. We use LLAMA-3.1-70b~\citep{grattafiori2024llama} to represent open-weights models without test-time reasoning. We use o3-mini~\citep{o3mini} as proprietary and DeepSeek-r1-70b~\citep{deepseekai2025deepseekr1incentivizingreasoningcapability} as open-weights models that support test-time reasoning. From a scientific point of view, DeepSeek-r1-70B is preferable, because we have access to the reasoning tokens. However, the costs of hosting DeepSeek-r1-70B locally are very high, because the reasoning processes can become very long, meaning that the outputs were several times larger than expected for only sorting. Though we cannot access the reasoning tokens, we also include the proprietary o3-mini model from OpenAI in our benchmark. The pricing for this model is reasonable and comparable to GPT-4o. Moreover, we can at least get access to the summary of the reasoning by using the ChatGPT web interface~(see Section~\ref{sec:results-o3-mini}). We did not include more reasoning models due to the prohibitive costs involved, e.g., OpenAIs o1, the more expensive, larger cousin o1 of the o3-mini model.

An important requirement for this model selection was that the whole sorting can happen within the context window of the model, which is the case for all of the above models. This avoids aspects like how longer contexts are supported to influence the results. Unfortunately, this is not always the case, as, e.g., the Gemma~\citep{team2024gemma}, Gemini~\citep{googleGeminiChat}, and Qwen 2.5~\citep{qwen2.5} model families tokenize numbers through their digits. Due to this, our advanced tasks with longer numbers were too long to fit within the context window. 

\section{Results}

Table~\ref{tbl:scores} shows the scores for the benchmark. Visualizations of the results for different list lengths for all tasks can be found in Appendix~\ref{sec:detailed-results}. We report the results of statistical tests conducted following the guidelines from \cite{benavoli2017time} for all list lengths in Appendix~\ref{sec:detailed-results}. All implementations we created for this work are publicly available online: BLINDED %\url{https://github.com/aieng-lab/sortbench}

\begin{table}
\centering
\begin{tabular}{llrrrr}
\toprule
& Model & $Model Score$ & $Sorting Score$ & $Faithfulness Score$ & $Validity Score$ \\
\midrule
\parbox[t]{2mm}{\multirow{7}{*}{\rotatebox[origin=c]{90}{Basic}}}
& o3-mini & \textbf{0.984} & 0.997 & \textbf{0.970} & \textbf{1.000} \\
& GPT-4o & 0.931 & 0.986 & 0.955 & 0.954 \\
& GPT-4o-mini & 0.901 & 0.978 & 0.888 & 0.964 \\
& Claude-3.5-Sonnet & 0.862 & \textbf{0.998} & 0.912 & 0.895 \\
& Claude-3.5-Haiku & 0.830 & 0.992 & 0.904 & 0.869 \\
& Llama-3.1 & 0.727 & 0.989 & 0.709 & 0.832 \\
& DeepSeek-r1 & 0.562 & 0.823 & 0.694 & 0.679 \\
\midrule
\parbox[t]{2mm}{\multirow{6}{*}{\rotatebox[origin=c]{90}{Advanced}}}
& o3-mini & \textbf{0.963} & \textbf{0.982} & \textbf{0.946} & \textbf{0.999} \\
& GPT-4o & 0.852 & 0.885 & 0.889 & 0.953 \\
& Claude-3.5-Sonnet & 0.775 & 0.925 & 0.870 & 0.854 \\
& GPT-4o-mini & 0.758 & 0.851 & 0.744 & 0.945 \\
& Claude-3.5-Haiku & 0.750 & 0.892 & 0.850 & 0.852 \\
& Llama-3.1 & 0.645 & 0.843 & 0.658 & 0.836 \\
& DeepSeek-r1 & 0.450 & 0.836 & 0.614 & 0.593 \\
\midrule
\parbox[t]{2mm}{\multirow{6}{*}{\rotatebox[origin=c]{90}{Debug}}}
& GPT-4o & \textbf{0.998} & 0.998 & \textbf{0.998} & \textbf{1.000} \\
& o3-mini & 0.995 & 0.999 & 0.991 & \textbf{1.000} \\
& GPT-4o-mini & 0.938 & 0.990 & 0.905 & 0.990 \\
& Claude-3.5-Sonnet & 0.863 & \textbf{1.000} & 0.913 & 0.895 \\
& Claude-3.5-Haiku & 0.844 & 0.998 & 0.898 & 0.886 \\
& Llama-3.1 & 0.780 & 0.980 & 0.803 & 0.858 \\
& DeepSeek-r1 & 0.615 & 0.991 & 0.756 & 0.684 \\
\midrule
\parbox[t]{2mm}{\multirow{7}{*}{\rotatebox[origin=c]{90}{All tasks}}}
& o3-mini & \textbf{0.977} & \textbf{0.990} & \textbf{0.965} & \textbf{0.999} \\
& GPT-4o & 0.914 & 0.940 & 0.937 & 0.969 \\
& GPT-4o-mini & 0.842 & 0.919 & 0.823 & 0.963 \\
& Claude-3.5-Sonnet & 0.819 & 0.962 & 0.891 & 0.875 \\
& Claude-3.5-Haiku & 0.795 & 0.944 & 0.875 & 0.866 \\
& Llama-3.1 & 0.704 & 0.915 & 0.716 & 0.843 \\
& DeepSeek-r1 & 0.524 & 0.887 & 0.677 & 0.638 \\
\bottomrule
\end{tabular}
\caption{Results sorted by the $ModelScore$, reported averaged over the different sets of of tasks (basic, advanced, debug) and overall.}
\label{tbl:scores}
\end{table}

\subsection{o3-mini performance dominates}
\label{sec:results-o3-mini}

The o3-mini test-time reasoning model from OpenAI outpeforms the other models in the benchmark: it almost always follows the instruction to output only a Python list, is most faithful to the input, and yields the best sorted list. For the $SortingScore$ the difference to Claude-3.5-Sonnet is extremely small, with a very small advantage for Claude-3.5-Sonnet for the basic and debug tasks and a small advantage for o3-mini on the advanced tasks. For all scores, GPT-4o is typically a bit worse than the o3-mini model, but the difference is not very large. In the debug tasks, GPT-4o even slightly outperforms o3-mini due to a better faithfulness. As we discuss in more detail in Section~\ref{sec:long-sequences}, the difference to the best other model (GPT-4o) is only significant for the longest lists with 256 items. 

Looking at the results in greater detail using the data reported in the appendix, o3-mini \textit{almost} solves the task at hand, i.e., is almost able to sort reliably. There are two open problems: First, o3-mini gets fooled by the NumberWords, where we observe problems with the $FaithfullnessScore$, while the $SortingScore$ remains very high. A look at the data shows the reason for this: the model often converts the strings into the corresponding integers and then sorts these. The output is, therefore, a sorted list but not with the expected, original list items. Another perspective is that the model got distracted from the task that only required looking at the syntax by instead using the semantics of the list items. We note that models without test-time reasoning do not have this problem. The second problem is also with the $FaithfullnessScore$, which drops for the longer lists of the Float-0:0001, ascii, AsCiI, and PrfxEnglish advanced tasks. The same happens for the English-Duplicate debug task. All these tasks share the common trait that they require a comparably many tokens for the list items. Thus, maintaining faithfulness to the input seems to become problematic as the context length increases.

To get insights into the reasoning of o3-mini, we used the ChatGPT Web frontend to sort a single list with 256 items for each task. While this does not give us access to the reasoning tokens, we can at least access a generated summary. The first reasoning step was always a repetition of the task at hand, that extended our prompt with a correct analysis of the data type and expected sorting order of the list. For about half of the tasks, it stopped there. In all these cases, the result was a correctly sorted list. In other cases, the model complained that using Python's \texttt{sorted} would be more convenient and that it needs to be ``extra careful''. We ran the NumberSorting two times: in one case, the model stuck to lexicographic sorting and ignored that the strings represented numbers, in the second case the model did was we described above, i.e., it identified that the strings represent numbers, converted them, and sorted them as integers. For one of the examples we tried, i.e., the basic English task, the reasoning process was very long, considered many individual words and characters, and (the summary provided on the Web page) even switched languages twice: the reasoning started in English, then switched to German, and then back to English. Whether these language switches are an artifact of the summarization or actually happened during the reasoning is unclear. Perhaps not surprisingly, this led to the worst performance among all lists we sorted using the Web-frontend. Overall, this qualitative analysis of the reasoning process is in line with the quantitative results from the benchmarks and shows that reasoning, in general, helps the model to elaborate on the prompt to make the result better, but also has an inherent risk that the model starts to \textit{overthink} causing hallucinations. While this did not happen very often with o3-mini, overthinking was a very large problem for DeepSeek-r1 (see Section \ref{sec:results-deepseek}). 

\subsection{Large, proprietary models are very good, even without test-time reasoning}

The other large, proprietary models in our benchmark, GPT-4o and Claude-3.5-Sonnet, also performs well and our statistical analysis shows that the difference to o3-mini only becomes relevant at list length of 128 for Claude-3-5-Sonnet and 256 for GPT-4o. 

GPT-4o is a close second to o3-mini and generally outperforms the other models, based on the $ModelScore$. It  consistently outperforms all models except o3-mini in terms of faithfulness, i.e., not changing the list items and also performs best in the advanced and debug tasks when it comes to the validity. Only for the basic task, the validity of GPT-4o-mini is slightly higher. This indicates that GPT-4o is very good at following the instruction that the output should only be a Python list and also that there is a relatively low rate of hallucinations in this setting, same as the reasoning counterpart o3-mini from OpenAI. The sorting capabilities of GPT-4o are also strong, though there are already a couple of mistakes for the longer lists in the basic task. The advanced tasks reveal a clear weakness for the sorting of strings in more complex setting (ascii, AsCiI, PrfxEnglish, and NumberWords), none of which works reliably for longer sequences. Moreover, the sorting of lists that include negative numbers (Int-n1000:1000) is also a problem for GPT-4o. The other advanced tasks, including the sorting of small ranges of floats, work well. The debug tasks show that the model also does not get easily tripped up by already sorted lists or duplicates within the lists. 

Claude-3.5-Sonnet is a bit better at sorting than GPT-4o, as shown by the better $SortingScore$ for all three groups of tasks. The sorting is perfect for the debug tasks and almost perfect for the basic tasks, i.e., the sorting capabilities are similar to those of o3-mini. For the advanced tasks, there are smaller problems for all tasks with longer list, but overall fewer bad sortings than with GPT-4o. Notably, the model handles both adversarial string tasks (PrfxEnglish, NumberWords) as well as lists with negative values (Int-n1000:1000) very well. However, the validity and faithfulness are both worse than for GPT-4o. The validity mostly suffers from not cleanly closing lists (see Section~\ref{sec:parsing-errors}). However, the lower faithfulness shows that the model has problems with keeping the list items as is. Claude-3.5-Sonnet has problems with the faithfulness for all string-based and float-based tasks, regardless of whether they are basic, advanced or for debugging. Only integers are reliably unchanged during the sorting. 

\subsection{Smaller proprietary models have the same properties as their large brothers}

The smaller variants of the proprietary models, GPT-4o-mini and Claude-3.5-Haiku have similar strengths and weakness than their larger counterparts, i.e., GPT-4o-mini is better when it comes to faithfulness and validity than Claude-3.5-Haiku, while Claude-3.5-Haiku has better $SortingScore$s. The gap between GPT-4o-mini and GPT-4o is fairly large for the advanced tasks, as both the sorting and faithfulness strongly suffer across all tasks. However, for the simpler basic and debug tasks, GPT-4o mini is performing very strongly, actually outperforming even the larger Claude-3.5-Sonnet model due to the higher validity. The gap between Claude-3.5-Sonnet and Claude-3.5-Haiku is more consistent, i.e., there is a relatively small gap across all metrics and task. 

\subsection{LLAMA has problems with long lists}

Overall, LLAMA-3.1 performs quite well in the benchmark: the performance is overall comparable to that of GPT-4o-mini and Claude-3.5-Haiku, which is expected given that we use the LLAMA variant with 70B parameters. When we analyze the performance of LLAMA in depth, we observe that long lists with 256 items are the reason, why the model does not score better. For all tasks, we observe a sharp drop in performance for these longer lists. This indicates that while the lists (incl. the sorted list) fit into the context window, the model obviously still has problems capturing relationships that span almost the whole context window. Further supporting this is that this drop in faithfulness and validity already happens earlier for the lists with floating point numbers and random strings as items: these lists require the most tokens among all list types we study, aggravating the problem. Moreover, LLAMA-3.1 has problems with negative numbers: here, the sign is often dropped when sorting, leading to a lack of faithfulness when sorting. 

\subsection{DeepSeek-r1 has problems with overthinking}
\label{sec:results-deepseek}

The clear disappointment in our experiment was the DeepSeek-r1 model. The problem can be easily summarized: through the reasoning the model got distracted. The initial reasoning steps were elaborating on the task (e.g., ``It seems like the input is a list of strings that should be sorted lexicographically''). When the reasoning stopped after this, the model performed the task as intended. Instead, the reasoning often continued for a long time, elaborating about unimportant details: in other words, the model was overthinking. While we have not checked all 1760 reasoning processes, we did a rough, qualitative analysis while studying parsing errors. Here, we observed three reasoning patterns that frequently emerged, that lead to different errors.

The first pattern was \textit{overthinking how sorting works}. For example, for strings the model looked at the individual letters of the first two list items, comparing each with the other, then for the next list item, etc. This is similar to the misbehavior we also observed in our smaller reasoning analysis with o3-mini. This extended the context to such a degree, that afterwards the general instruction (sorting of the list as Python) was lost, and the model output something generic that describes how sorting works with the words from the start of the list as example. 

The second pattern was \textit{looking for relationships between items}. This typically happened for numbers, where instead of just sorting the list, DeepSeek-r1 started to hallucinate and tried to find an equation that describes the list as a sequence. As a result, the output was not a list anymore, but rather some description of the pattern or something that should describe the relationship. Another example was that the model converted the strings of the NumberWords into actual numbers (breaking the list format) or instead of sorting the NumberWords lexicographically, they were instead sorted by the numeric values they represent. Notably, this even happened when the reasoning process initially correctly stated that the list is with strings and should be sorted lexicographically according to the task. This is another aspect we also observed with o3-mini. 

The third pattern was the \textit{list repetition}, that we also worried about initially: the reasoning process repeated the lists many times, and tried to check the sorting error in between. Through this, the result did not get better, but rather worse over time: with every iteration of the reasoning process, this added a risk of losing items or breaking the format, likelihood of a valid and faithful output. This pattern was not observed for o3-mini, which only repeated short subsequences of lists during the reasoning -- though this may still have happened, and we just did not see this in the summaries that we had access to. 

\begin{figure}
\centering
\includegraphics[width=0.5\linewidth]{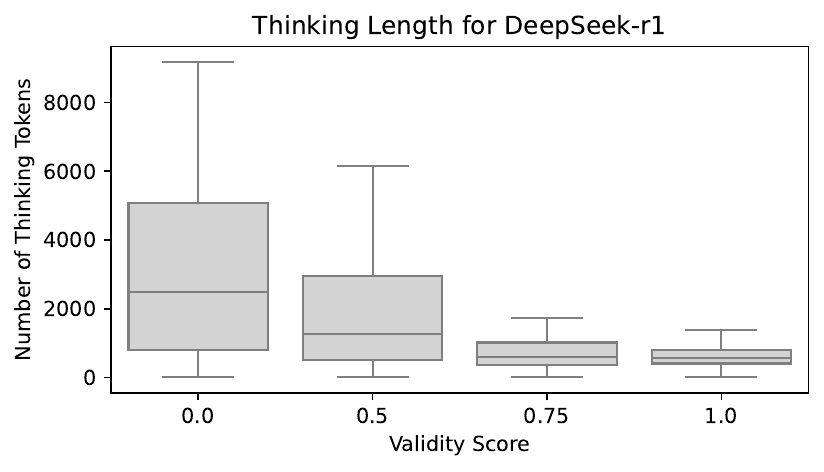}
\caption{Boxplot of the  number of tokens during the reasoning process for the different values of the $ValidityScore$.}
\label{fig:thinking-tokens}
\end{figure}

For all these patterns, it holds that the model generates more reasoning tokens than necessary for the task. This pushes the initial instruction from our prompt further away from the output, which often leads to hallucinations what the actual task is, resulting in those hallucinations that we described in the the second pattern or instead just describing how sorting works. The data presented in Figure~\ref{fig:thinking-tokens} confirms this: the mean number of reasoning tokens is by far the longest for outputs we could not parse, followed by cases that we could only parse with corner case handling. The cases that are almost valid or completely valid Python lists have by far the shortest reasoning processes. 

The above observations explain the overall weak performance of DeepSeek-r1, which is driven by issues with the validity and faithfulness of the output, which just got lost in the reasoning process. However, if the output was indeed a list, the sorting was good (though still worse than for the large, proprietary models). 

\subsection{Long sequences are a problem}
\label{sec:long-sequences}

Across all tasks and models, the $SortBenchScore$ goes down with longer sequences. Moreover, when we consider the statistical analysis of the results for different list length (see Appendix~\ref{sec:details-stats}), we find that the differences between LLMs are insignificant for lists with at most 8 items. With 16 items, we start to observe a gap between the three best models (o3-mini, GPT-4o, and Claude-3.5-Sonnet) and the others. At this length, the overthinking problems of DeepSeek-r1 also start to manifest. At a length of 128, we still cannot measure a significant difference between o3-mini and GPT-4o, though Claude-3.5-Sonnet is different with a large effect size from the others. Only at length 256 all differences between the models become significant, except for Claude-3.5-Sonnet and Claude-3.5-Haiku. While the lack of a difference at this length between the two Claude variants may seem surprising, this is a consequence of the relatively small difference in sorting capability and the similar problems with producing valid outputs for list of this lengths that both models share. 

When we analyze these trends over the $SortingScore$, $FaithfulnessScore$, and $ValidityScore$, we see that this mostly holds: having more list items increase the risk of invalid outputs, lead to the LLM being less faithful to the original list, and reduce the quality of the sorting. We say mostly, because some models actually show an increase in the $SortingScore$ for longer lists, e.g., DeepSeek-r1 for the advanced tasks. However, this does not really mean that the models get better at sorting with longer list. A careful look at the data reveals that this increase is accompanied by a sharp drop in the $FaithfulnessScore$, i.e., the LLMs just output some sorted output, instead of solving the actual task in these cases. Overall, this demonstrates quite clearly that long contexts are still a problem for current LLMs, while all LLMs are fairly good at dealing with short contexts. 

\subsection{Reasons for parsing errors}
\label{sec:parsing-errors}

\begin{figure}
    \centering
\includegraphics[width=0.7\linewidth]{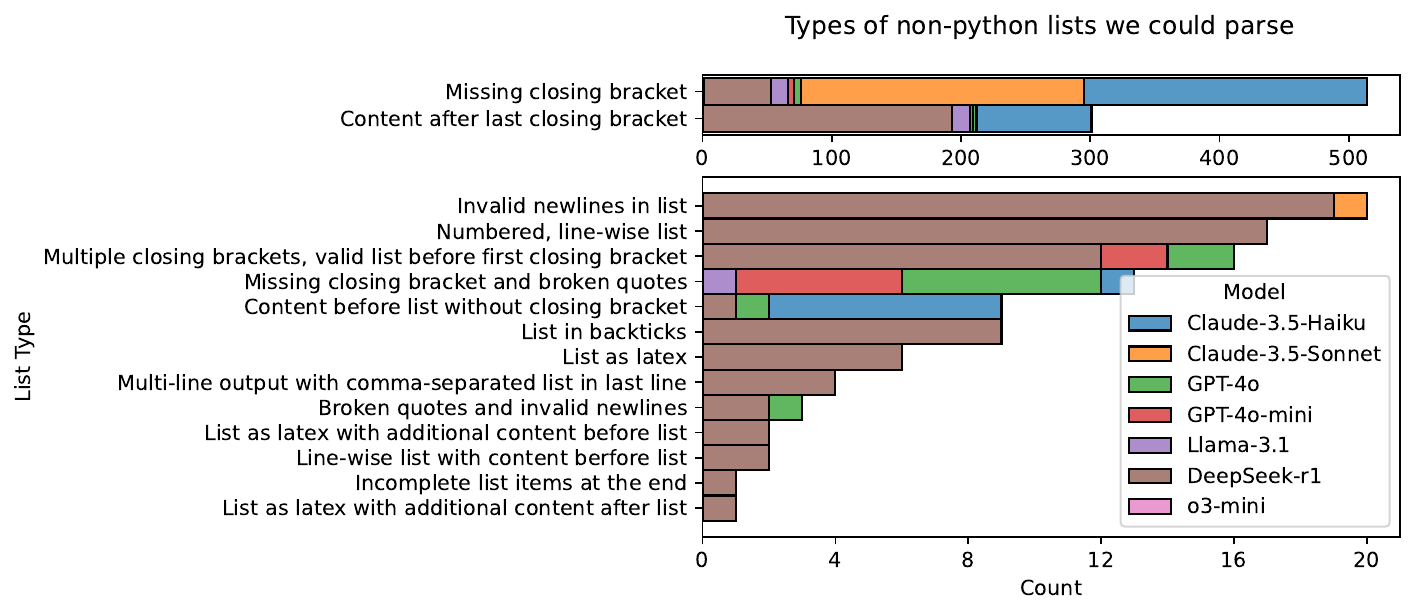}
    \caption{Different types of lists we found in the output of the LLMs that were not valid Python. The plot is split, since the first two types appear hundreds of times, while the others are rather corner cases with at most twenty instances.}
    \label{fig:error_types}
\end{figure}

When we introduced the $ValidityScore$, we already stated that we observed many types of sorted lists, that were not the required Python lists. Figure \ref{fig:error_types} shows the errors we observed. In total, we defined fifteen customized list parsers to handle lists that were not valid Python. The most common problem were missing closing brackets, with about 200 instances for both Claude variants, and, additionally, the other models where this rather happened as a corner case. We also observed a variant of this where a missing closing bracket was combined with the quotes of a string list being broken (e.g., because there was no closing quote for the last list item). The other case that was driving most parsing errors is that there was content after the list. Such content was always an explanation for the sorting or the task that was performed. This happened very often with DeepSeek-r1, but was also not uncommon with Claude-3.5-Haiku. In general, content before or after the list was very often the problem with DeepSeek-r1, often also combined with other list types, including line-wise list (one item per line), enumerations (one item per line starting with the list index), lists in latex formats (e.g., \textbackslash{}boxed or in math-mode), and in backticks to indicate code in MarkDown. Overall, ``creatively'' is the best description for how DeepSeek-r1 determined the output format. 

\section{Discussion}

\subsection{SortBench Basic, Advanced, and Debug help us to understand LLMs}

Our results show that SortBench is a suitable way to understand core properties of LLMs: Are they able to follow simple instructions, even if the output that is generated is long? Do the embeddings encode basic properties of tokens like their order in relation to items of the same type? In summary, our results show that test-time reasoning improves these capabilities (as expected), but comes with a risk of overthinking which may decreases the performance. Here, we observe a very large difference between the two test-time reasoning models in our benchmark: o3-mini has the overthinking mostly under control, while it is rampant for DeepSeek-r1. This shows both the potential, but also the risk of this sort of LLM training. Further, we showed that all LLMs, regardless of their size, can in principle sort lists (i.e., have this information in their embeddings), but that size gets important for longer lists. Interestingly, this means that even such a basic property (order) is not captured reliably over longer contexts, without huge models. Finally, this simple task demonstrates that you cannot trust LLMs to follow instructions. Even the (supposedly) simple debugging task of ``sorting'' an already sorted lists is typically not solved perfectly, due to the inherently random nature of LLMs. 

\subsection{Towards SortBench Hard}

The sorting tasks we have considered here are still rather simple, yet sufficient to bring even current state-of-the-art models to their limits. However, o3-mini is close to solving this benchmark. Still, sorting offers many more options that can be used to test models, including multi-modal models that include vision. There are several such tasks that we foresee for a future hard version of the SortBench:
\begin{itemize}
    \item Consideration of sorting orders (i.e. ascending and descending).
    \item Other ways to express numbers, e.g., as fractions or in the scientific exponential notation. 
    \item Sorting within other languages than English.
    \item Sorting of long strings (e.g., sentences).
    \item Sorting with non-ASCII character sets.
    \item Sorting of well-known categories, such as cloth sizes (XS, S, M).
    \item Sorting of manually defined categories, i.e., the sorting order is defined as part of a prompt. 
    \item Sorting of complex data types with multiple attributes for sorting (e.g., sorting two-column tabular data, first by one column, in case of ties by the second column.
    \item Sorting complex data with different sorting orders for different attributes.
    \item Sorting of items in images by their size (absolute in pixels).
    \item Sorting of items in images relative to a reference item (e.g., size in relation to a banana displayed in the image). 
    \item Sorting of strings by the length of the strings.
    \item Sorting strings by their edit-distance to a baseline.
    \item Sorting items by their distance in an ontology.
\end{itemize}

As can be seen, this problem can easily be extended to require more complex skills, e.g., being able to apply multiple sorting orders at the same time, understanding manually defined sorting orders, and being able to develop suitable comparison mechanisms for complex relationships. 

\subsection{Avoiding overfitting with SortBench}

An advantage of a benchmark like SortBench is that new data can be generated easily. This means that regular releases of this benchmark with new data, that was not yet seen by the models beforehand, are possible. We plan yearly releases of SortBench to prevent overfitting, which can easily occur if this data becomes part of the training of models. This is in stark contrast to commonly used benchmarks like MMLU~\citep{hendrycks2020measuring} which require a large amount of effort to create, update, and replace or the Chatbot Arena~\citep{chiang2024chatbot} that requires users to constantly test and compare models. 

\subsection{Limitations}

While SortBench provides good results and seems like a simple test that is yet capable of testing LLMs and bringing them to their limits, one may criticize whether the benchmark is actually useful, as it seemingly does not test any direct real-world use case, since no one would actually start sorting items this way. However, especially users without deep knowledge about how LLMs work or the domain in which they are using them, are often not aware of LLMs limitations. For example, users may use them to analyze tabular data, e.g., spreadsheets~\citep{chen2024sheetagent}. For such tasks, a basic understanding of orders, as well as the capability to faithfully handle long context-information is crucial. Moreover, sorting also reveals interesting information about what is actually contained in the embeddings, i.e., whether the embeddings encode information like lexical or numerical order reliably. Thus, we believe that SortBench is a suitable benchmark for LLMs that is easy to run, scale, and make more difficult for future LLM generations. 

\section{Conclusion}

Sorting is a basic skill that humans learn at an early age. The SortBench benchmark tests the capabilities of LLMs at this task in different scenarios. We find that even current test-time reasoning models struggle at this task for longer lists, e.g., of random strings or very small floating point numbers. Moreover, test-time reasoning models tend to overthink and, e.g., ignore lexicographic orders if the strings represents numbers. SortBench further shows that all current models have problems with not changing the longer lists that should be sorted, an aspect we refer to as faithfulness. Overall, SortBench is a suitable benchmark to rank LLMs by their capability to follow instructions, handle inputs of different lengths, and solve a task that requires basic reasoning capabilities.

%\subsubsection*{Broader Impact Statement}
%In this optional section, TMLR encourages authors to discuss possible repercussions of their work,
%notably any potential negative impact that a user of this research should be aware of. 
%Authors should consult the TMLR Ethics Guidelines available on the TMLR website
%for guidance on how to approach this subject.

\bibliography{tmlr}

\begin{thebibliography}{23}
\providecommand{\natexlab}[1]{#1}
\providecommand{\url}[1]{\texttt{#1}}
\expandafter\ifx\csname urlstyle\endcsname\relax
  \providecommand{\doi}[1]{doi: #1}\else
  \providecommand{\doi}{doi: \begingroup \urlstyle{rm}\Url}\fi

\bibitem[Anthropic(2024)]{anthropicIntroducingClaude}
Anthropic.
\newblock {I}ntroducing {C}laude 3.5 {S}onnet --- anthropic.com.
\newblock \url{https://www.anthropic.com/news/claude-3-5-sonnet}, 2024.
\newblock [Accessed 27-03-2025].

\bibitem[{AoPS Online}(2024)]{artofproblemsolvingProblemSolving}
{AoPS Online}.
\newblock {A}rt of {P}roblem {S}olving --- artofproblemsolving.com.
\newblock \url{https://artofproblemsolving.com/wiki/index.php/AIME_Problems_and_Solutions}, 2024.
\newblock [Accessed 27-03-2025].

\bibitem[Benavoli et~al.(2014)Benavoli, Corani, Mangili, Zaffalon, and Ruggeri]{benavoli2014bayesian}
Alessio Benavoli, Giorgio Corani, Francesca Mangili, Marco Zaffalon, and Fabrizio Ruggeri.
\newblock A bayesian wilcoxon signed-rank test based on the dirichlet process.
\newblock In \emph{International conference on machine learning}, pp.\  1026--1034, 2014.

\bibitem[Benavoli et~al.(2017)Benavoli, Corani, Dem{\v{s}}ar, and Zaffalon]{benavoli2017time}
Alessio Benavoli, Giorgio Corani, Janez Dem{\v{s}}ar, and Marco Zaffalon.
\newblock Time for a change: a tutorial for comparing multiple classifiers through bayesian analysis.
\newblock \emph{The Journal of Machine Learning Research}, 18\penalty0 (1):\penalty0 2653--2688, 2017.

\bibitem[Chen et~al.(2024)Chen, Yuan, Zhang, Zheng, Liu, Ni, and Hao]{chen2024sheetagent}
Yibin Chen, Yifu Yuan, Zeyu Zhang, Yan Zheng, Jinyi Liu, Fei Ni, and Jianye Hao.
\newblock Sheetagent: A generalist agent for spreadsheet reasoning and manipulation via large language models.
\newblock In \emph{ICML 2024 Workshop on LLMs and Cognition}, 2024.

\bibitem[Chiang et~al.(2024)Chiang, Zheng, Sheng, Angelopoulos, Li, Li, Zhu, Zhang, Jordan, Gonzalez, et~al.]{chiang2024chatbot}
Wei-Lin Chiang, Lianmin Zheng, Ying Sheng, Anastasios~Nikolas Angelopoulos, Tianle Li, Dacheng Li, Banghua Zhu, Hao Zhang, Michael Jordan, Joseph~E Gonzalez, et~al.
\newblock Chatbot arena: An open platform for evaluating llms by human preference.
\newblock In \emph{Forty-first International Conference on Machine Learning}, 2024.

\bibitem[Cohen(2013)]{cohen2013statistical}
Jacob Cohen.
\newblock \emph{Statistical power analysis for the behavioral sciences}.
\newblock Academic press, 2013.

\bibitem[DeepSeek-AI(2025)]{deepseekai2025deepseekr1incentivizingreasoningcapability}
DeepSeek-AI.
\newblock Deepseek-r1: Incentivizing reasoning capability in llms via reinforcement learning, 2025.
\newblock URL \url{https://arxiv.org/abs/2501.12948}.

\bibitem[Dem{\v{s}}ar(2006)]{demvsar2006statistical}
Janez Dem{\v{s}}ar.
\newblock Statistical comparisons of classifiers over multiple data sets.
\newblock \emph{Journal of Machine learning research}, 7\penalty0 (Jan):\penalty0 1--30, 2006.

\bibitem[Fellbaum(2010)]{fellbaum2010wordnet}
Christiane Fellbaum.
\newblock Wordnet.
\newblock In \emph{Theory and applications of ontology: computer applications}, pp.\  231--243. Springer, 2010.

\bibitem[{Gemma Team} et~al.(2024){Gemma Team}, Riviere, Pathak, Sessa, Hardin, Bhupatiraju, Hussenot, Mesnard, Shahriari, Ram{\'e}, et~al.]{team2024gemma}
{Gemma Team}, Morgane Riviere, Shreya Pathak, Pier~Giuseppe Sessa, Cassidy Hardin, Surya Bhupatiraju, L{\'e}onard Hussenot, Thomas Mesnard, Bobak Shahriari, Alexandre Ram{\'e}, et~al.
\newblock Gemma 2: Improving open language models at a practical size.
\newblock \emph{arXiv preprint arXiv:2408.00118}, 2024.

\bibitem[Google(2024)]{googleGeminiChat}
Google.
\newblock {G}emini - chat to supercharge your ideas.
\newblock \url{https://gemini.google.com/}, 2024.
\newblock [Accessed 27-03-2025].

\bibitem[Grattafiori et~al.(2024)Grattafiori, Dubey, Jauhri, Pandey, Kadian, Al-Dahle, Letman, Mathur, Schelten, Vaughan, et~al.]{grattafiori2024llama}
Aaron Grattafiori, Abhimanyu Dubey, Abhinav Jauhri, Abhinav Pandey, Abhishek Kadian, Ahmad Al-Dahle, Aiesha Letman, Akhil Mathur, Alan Schelten, Alex Vaughan, et~al.
\newblock The llama 3 herd of models.
\newblock \emph{arXiv preprint arXiv:2407.21783}, 2024.

\bibitem[Hendrycks et~al.(2020)Hendrycks, Burns, Basart, Zou, Mazeika, Song, and Steinhardt]{hendrycks2020measuring}
Dan Hendrycks, Collin Burns, Steven Basart, Andy Zou, Mantas Mazeika, Dawn Song, and Jacob Steinhardt.
\newblock Measuring massive multitask language understanding.
\newblock \emph{arXiv preprint arXiv:2009.03300}, 2020.

\bibitem[Hurst et~al.(2024)Hurst, Lerer, Goucher, Perelman, Ramesh, Clark, Ostrow, Welihinda, Hayes, Radford, et~al.]{hurst2024gpt}
Aaron Hurst, Adam Lerer, Adam~P Goucher, Adam Perelman, Aditya Ramesh, Aidan Clark, AJ~Ostrow, Akila Welihinda, Alan Hayes, Alec Radford, et~al.
\newblock Gpt-4o system card.
\newblock \emph{arXiv preprint arXiv:2410.21276}, 2024.

\bibitem[Jaech et~al.(2024)Jaech, Kalai, Lerer, Richardson, El-Kishky, Low, Helyar, Madry, Beutel, Carney, et~al.]{jaech2024openai}
Aaron Jaech, Adam Kalai, Adam Lerer, Adam Richardson, Ahmed El-Kishky, Aiden Low, Alec Helyar, Aleksander Madry, Alex Beutel, Alex Carney, et~al.
\newblock Openai o1 system card.
\newblock \emph{arXiv preprint arXiv:2412.16720}, 2024.

\bibitem[Kruschke \& Liddell(2018)Kruschke and Liddell]{kruschke2018bayesian}
John~K Kruschke and Torrin~M Liddell.
\newblock The bayesian new statistics: Hypothesis testing, estimation, meta-analysis, and power analysis from a bayesian perspective.
\newblock \emph{Psychonomic Bulletin \& Review}, 25\penalty0 (1):\penalty0 178--206, 2018.

\bibitem[{OpenAI}(2025)]{o3mini}
{OpenAI}.
\newblock o3-mini system card.
\newblock \url{https://openai.com/index/o3-mini-system-card/}, 2025.
\newblock [Accessed 27-03-2025].

\bibitem[Price(2024)]{arcprizePrize}
ARC Price.
\newblock {A}{R}{C} {P}rize --- arcprize.org.
\newblock \url{https://arcprize.org/}, 2024.
\newblock [Accessed 27-03-2025].

\bibitem[Radford et~al.(2018)Radford, Narasimhan, Salimans, Sutskever, et~al.]{radford2018improving}
Alec Radford, Karthik Narasimhan, Tim Salimans, Ilya Sutskever, et~al.
\newblock Improving language understanding by generative pre-training.
\newblock 2018.

\bibitem[Sennrich et~al.(2015)Sennrich, Haddow, and Birch]{sennrich2015neural}
Rico Sennrich, Barry Haddow, and Alexandra Birch.
\newblock Neural machine translation of rare words with subword units.
\newblock \emph{arXiv preprint arXiv:1508.07909}, 2015.

\bibitem[Yang et~al.(2024)Yang, Yang, Zhang, Hui, Zheng, Yu, Li, Liu, Huang, Wei, Lin, Yang, Tu, Zhang, Yang, Yang, Zhou, Lin, Dang, Lu, Bao, Yang, Yu, Li, Xue, Zhang, Zhu, Men, Lin, Li, Xia, Ren, Ren, Fan, Su, Zhang, Wan, Liu, Cui, Zhang, and Qiu]{qwen2.5}
An~Yang, Baosong Yang, Beichen Zhang, Binyuan Hui, Bo~Zheng, Bowen Yu, Chengyuan Li, Dayiheng Liu, Fei Huang, Haoran Wei, Huan Lin, Jian Yang, Jianhong Tu, Jianwei Zhang, Jianxin Yang, Jiaxi Yang, Jingren Zhou, Junyang Lin, Kai Dang, Keming Lu, Keqin Bao, Kexin Yang, Le~Yu, Mei Li, Mingfeng Xue, Pei Zhang, Qin Zhu, Rui Men, Runji Lin, Tianhao Li, Tingyu Xia, Xingzhang Ren, Xuancheng Ren, Yang Fan, Yang Su, Yichang Zhang, Yu~Wan, Yuqiong Liu, Zeyu Cui, Zhenru Zhang, and Zihan Qiu.
\newblock Qwen2.5 technical report.
\newblock \emph{arXiv preprint arXiv:2412.15115}, 2024.

\bibitem[Zhang et~al.(2023)Zhang, Dong, Li, Zhang, Sun, Wang, Li, Hu, Zhang, Wu, et~al.]{zhang2023instruction}
Shengyu Zhang, Linfeng Dong, Xiaoya Li, Sen Zhang, Xiaofei Sun, Shuhe Wang, Jiwei Li, Runyi Hu, Tianwei Zhang, Fei Wu, et~al.
\newblock Instruction tuning for large language models: A survey.
\newblock \emph{arXiv preprint arXiv:2308.10792}, 2023.

\end{thebibliography}
\bibliographystyle{tmlr}

\appendix
\section{Appendix}

This appendix provides additional details regarding our analysis with visualizations in Appendix~\ref{sec:detailed-results} and statistical analysis in Appendix~\ref{sec:details-stats}.

\FloatBarrier

\subsection{Visualization of results}
\label{sec:detailed-results}

\begin{figure}[h]
\centering
\includegraphics[width=0.7\linewidth]{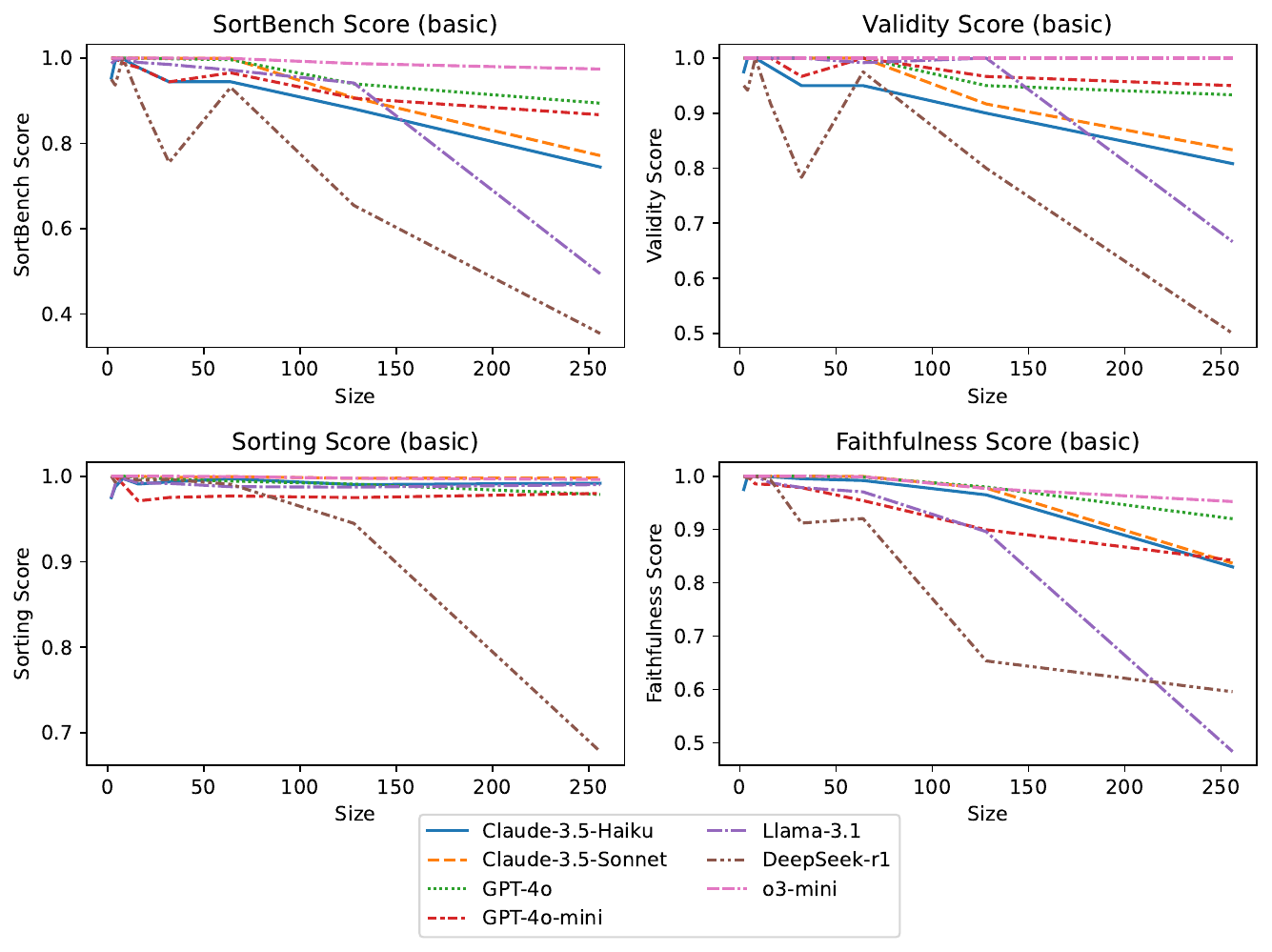}
\caption{Aggregated results for all basic tasks by list size}
\label{fig:results-basic-aggr}
\end{figure}

\begin{figure}[h]
\centering
\includegraphics[width=0.6\linewidth]{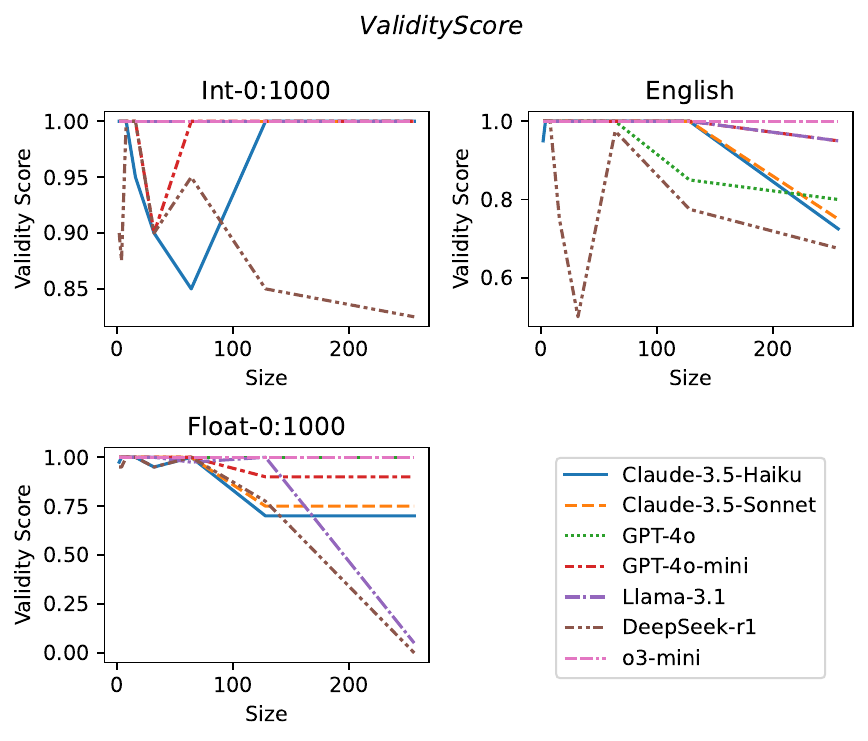}
\caption{$ValidityScore$ for all basic tasks by list size}
\label{fig:results-basic-val}
\end{figure}

\begin{figure}[h]
\centering
\includegraphics[width=0.6\linewidth]{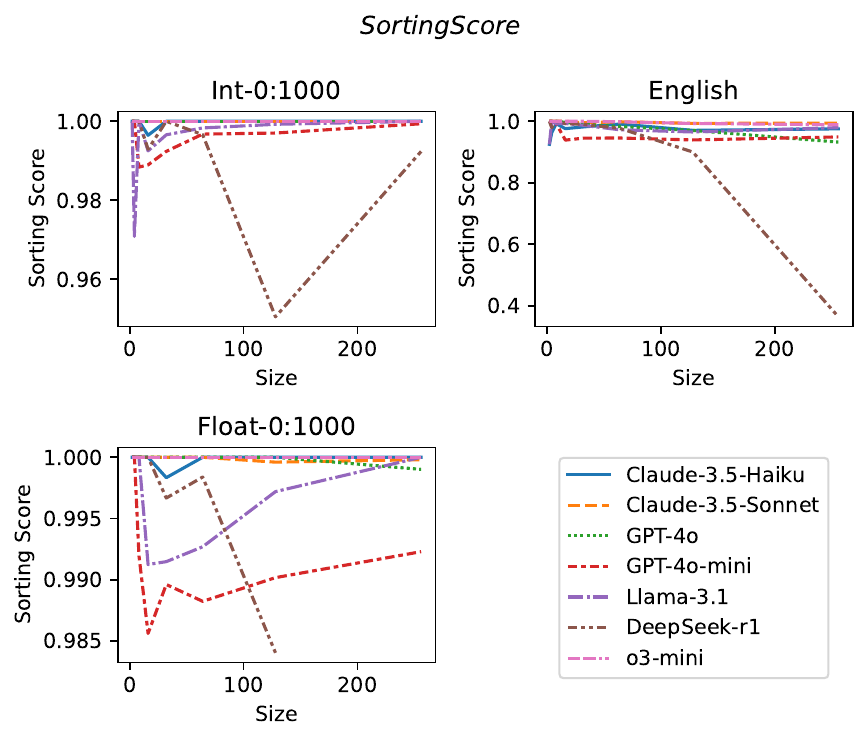}
\caption{$SortingScore$ for all basic tasks by list size}
\label{fig:results-basic-sort}
\end{figure}

\begin{figure}[h]
\centering
\includegraphics[width=0.6\linewidth]{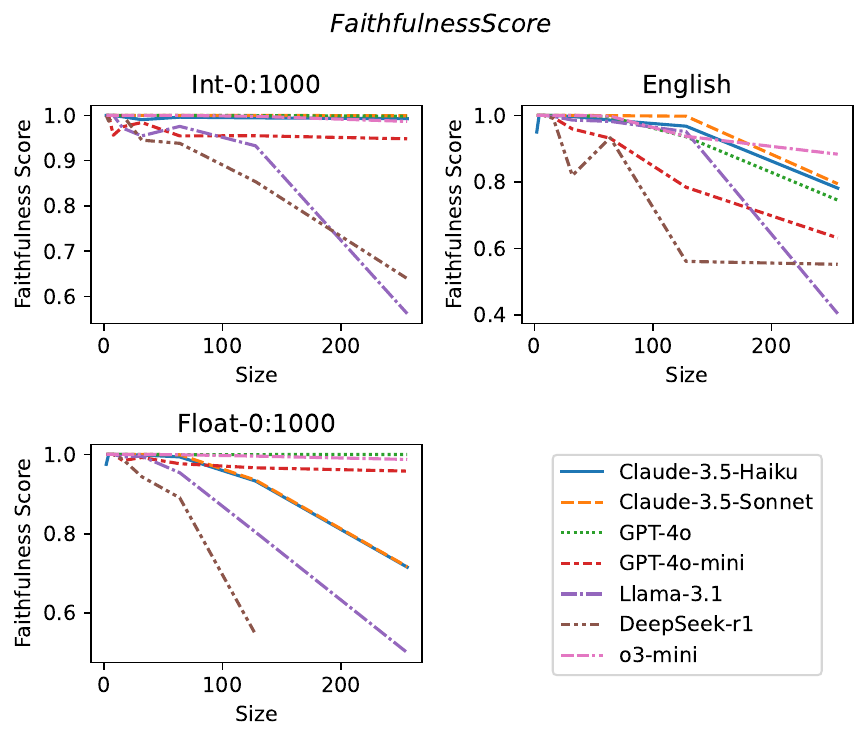}
\caption{$FaithfulnessScore$ for all basic tasks by list size}
\label{fig:results-basic-faith}
\end{figure}

\begin{figure}[h]
\centering
\includegraphics[width=0.7\linewidth]{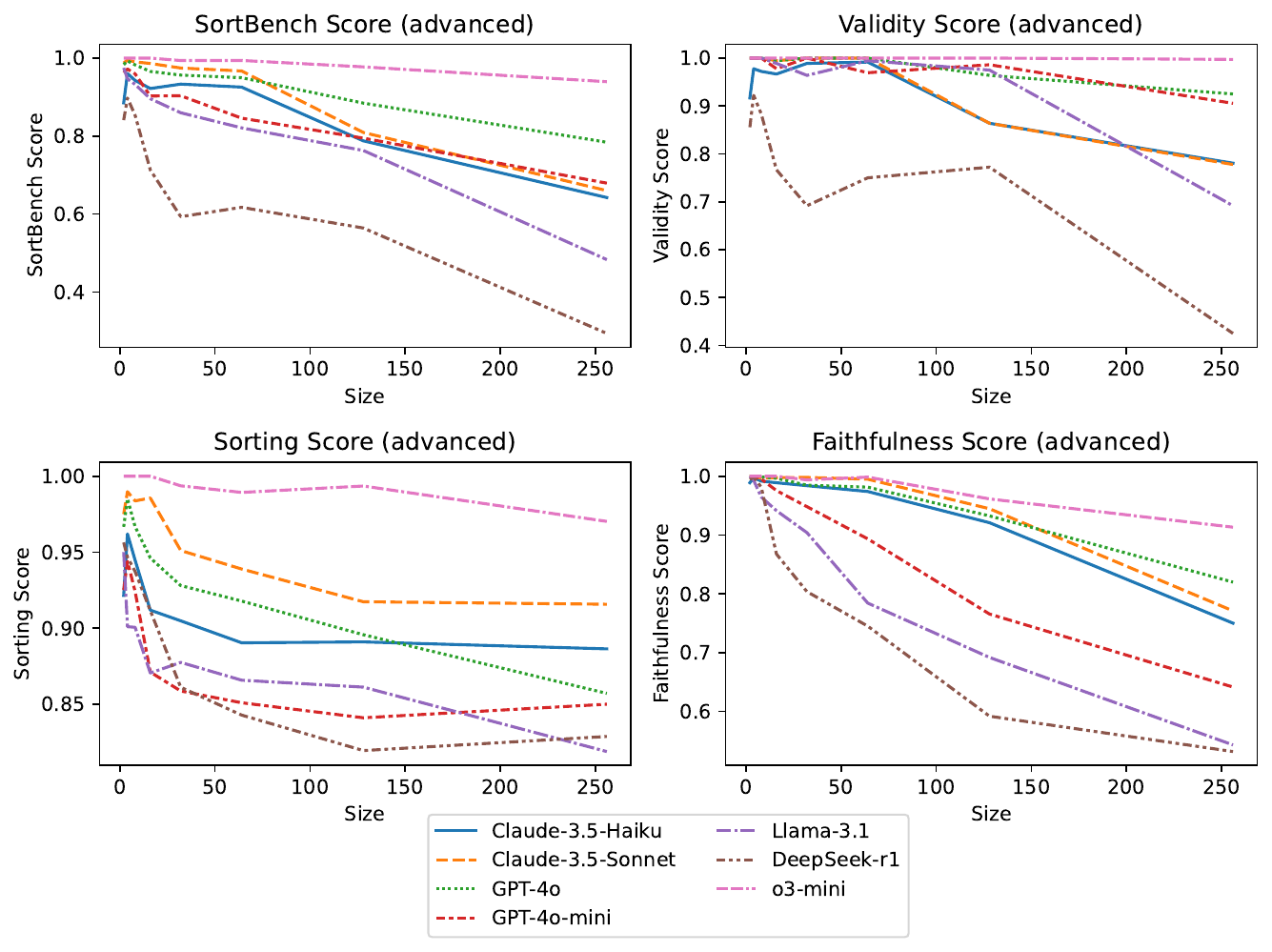}
\caption{Aggregated results for all advanced tasks by list size}
\label{fig:results-advanced-aggr}
\end{figure}

\begin{figure}[h]
\centering
\includegraphics[width=0.6\linewidth]{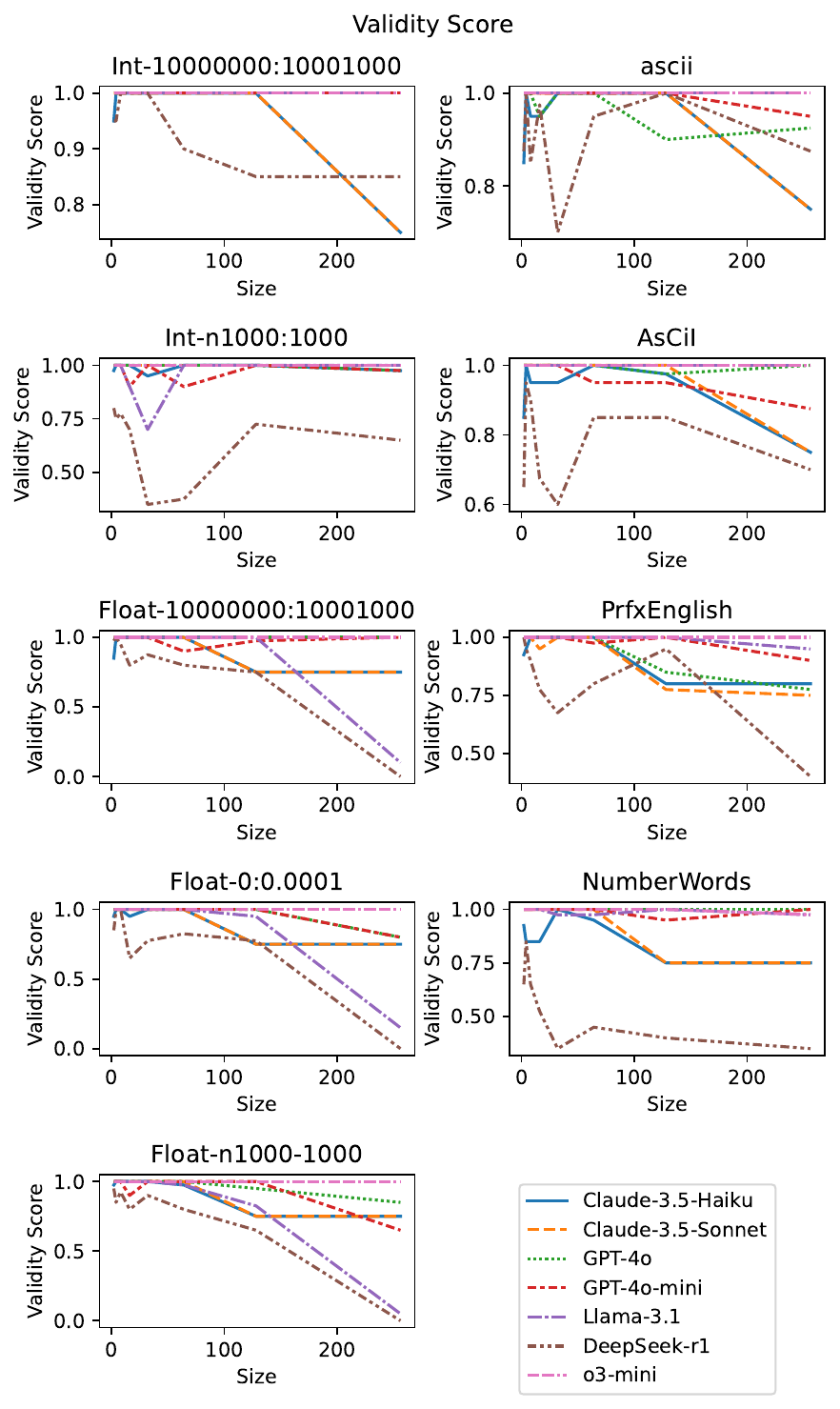}
\caption{$ValidityScore$ for all advanced tasks by list size}
\label{fig:results-advanced-val}
\end{figure}

\begin{figure}[h]
\centering
\includegraphics[width=0.6\linewidth]{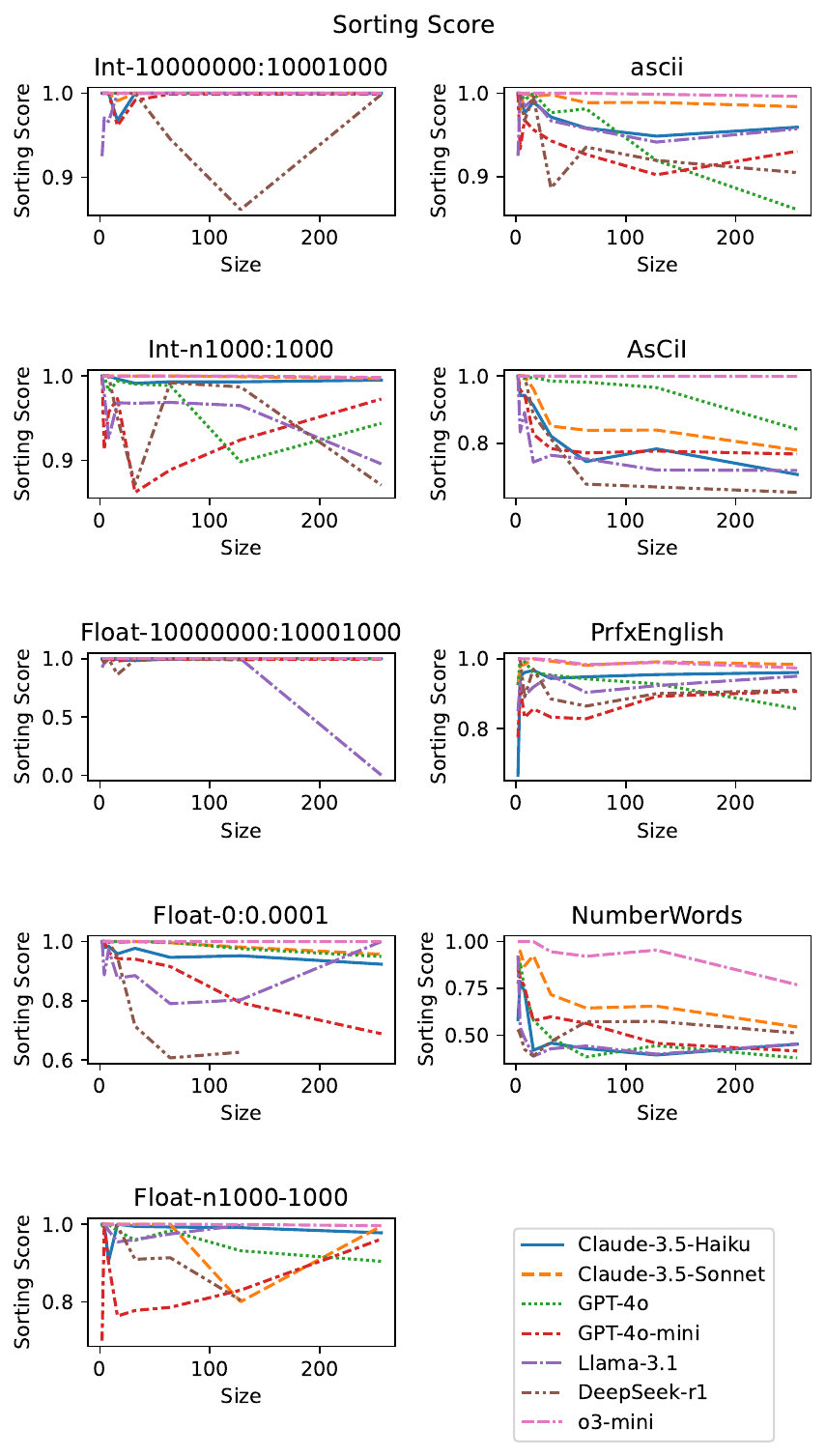}
\caption{$SortingScore$ for all advanced tasks by list size}
\label{fig:results-advanced-sort}
\end{figure}

\begin{figure}[h]
\centering
\includegraphics[width=0.6\linewidth]{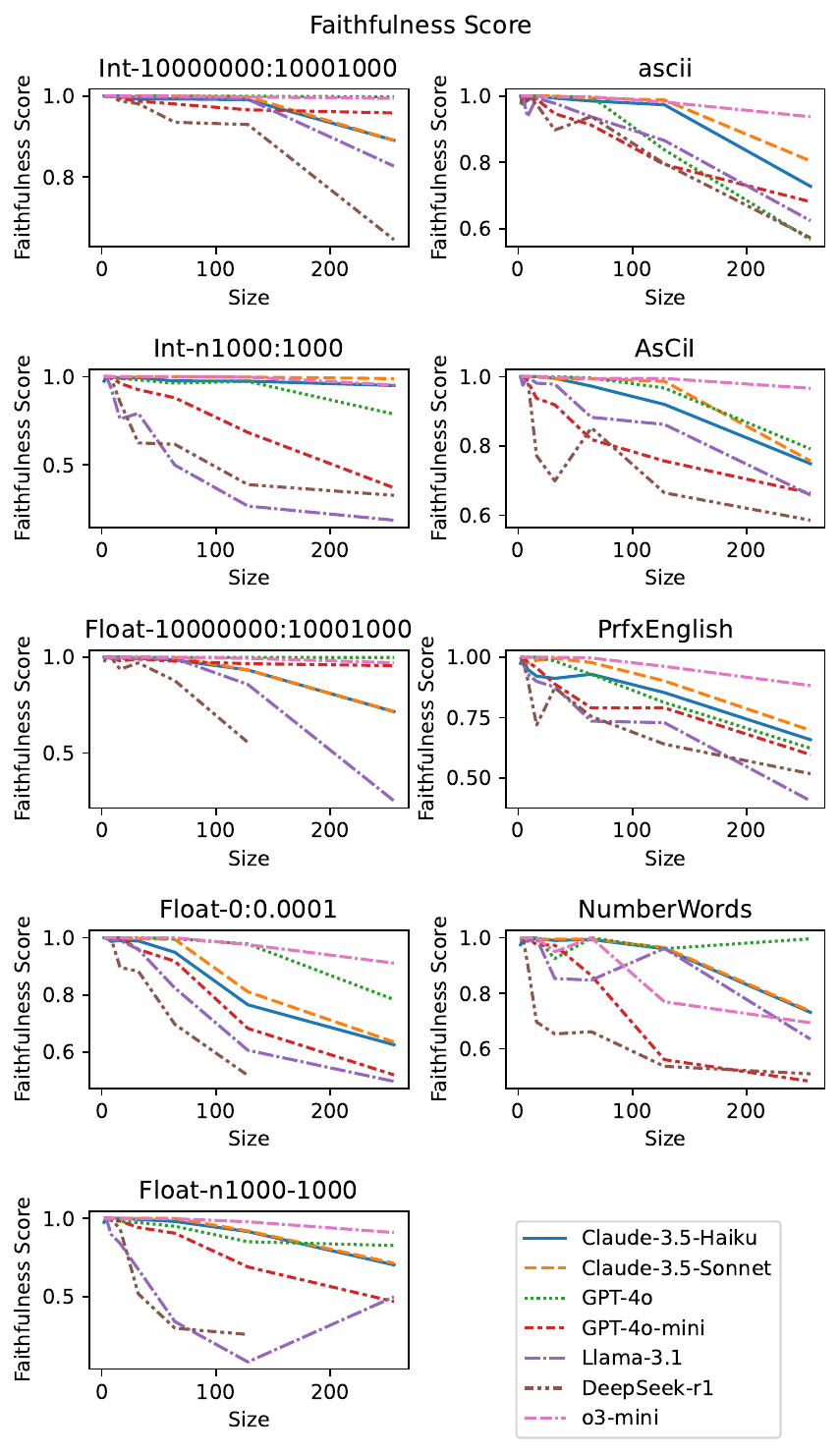}
\caption{$FaithfulnessScore$ for all advanced tasks by list size}
\label{fig:results-advanced-faith}
\end{figure}

\begin{figure}[h]
\centering
\includegraphics[width=0.7\linewidth]{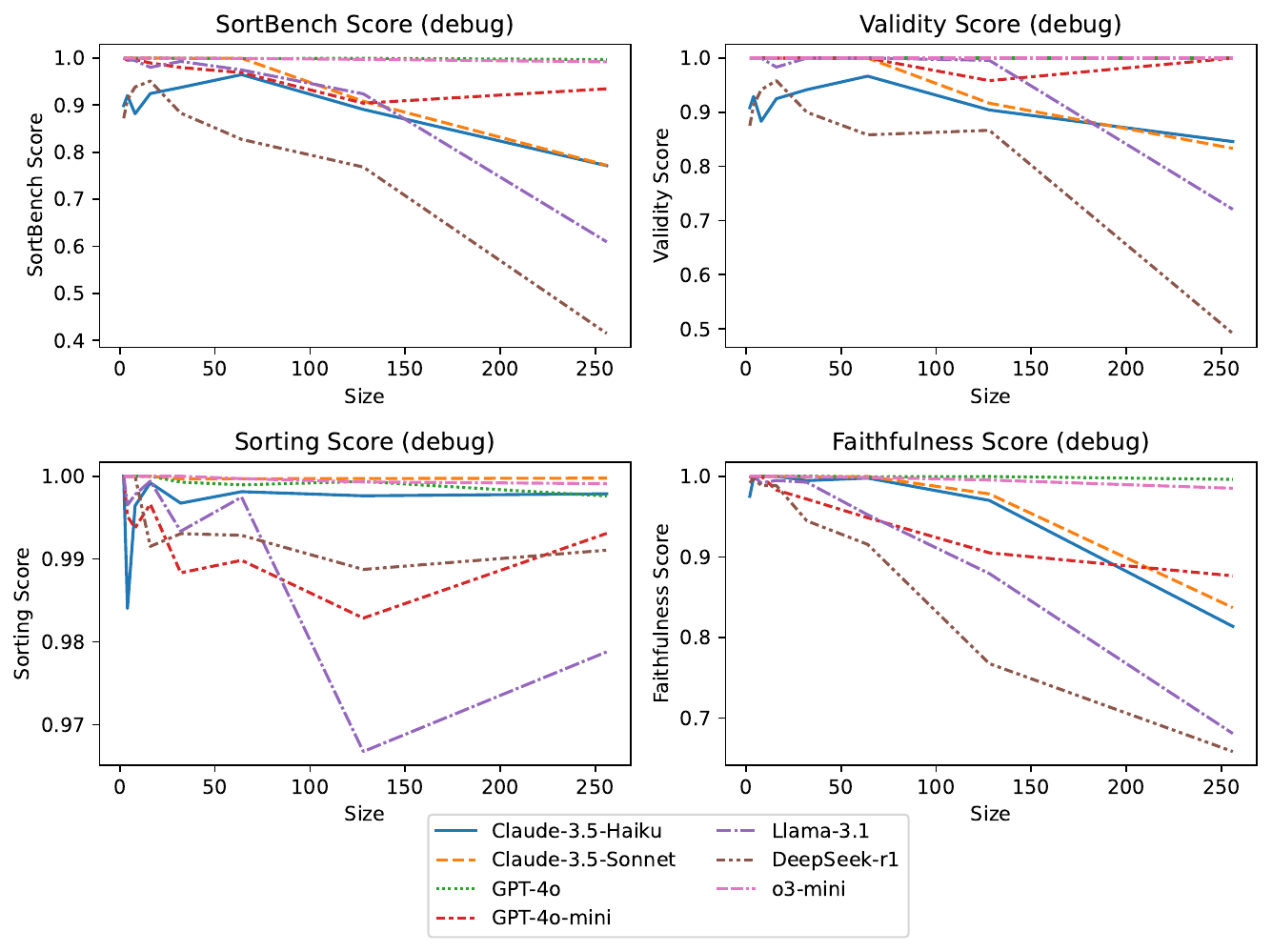}
\caption{Aggregated results for all debug tasks by list size}
\label{fig:results-debug-aggr}
\end{figure}

\begin{figure}[h]
\centering
\includegraphics[width=0.6\linewidth]{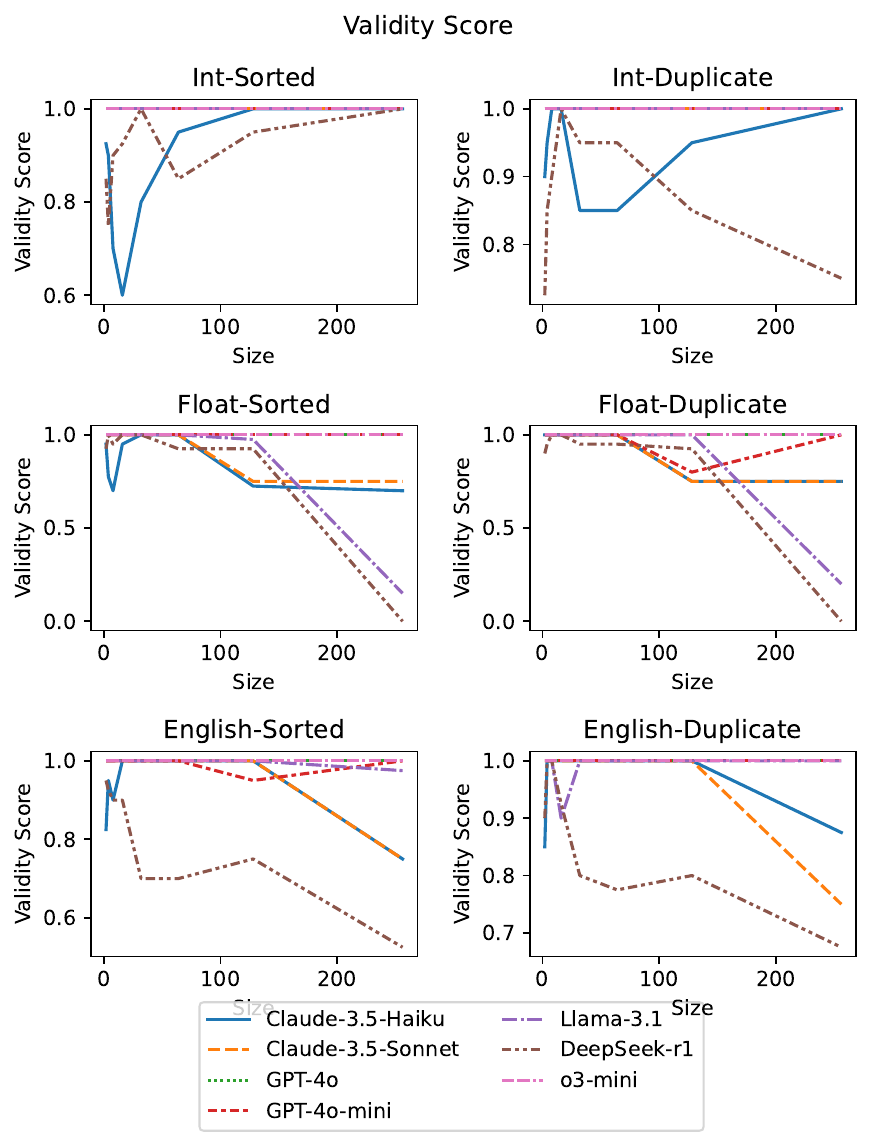}
\caption{$ValidityScore$ for all debug tasks by list size}
\label{fig:results-debug-val}
\end{figure}

\begin{figure}[h]
\centering
\includegraphics[width=0.6\linewidth]{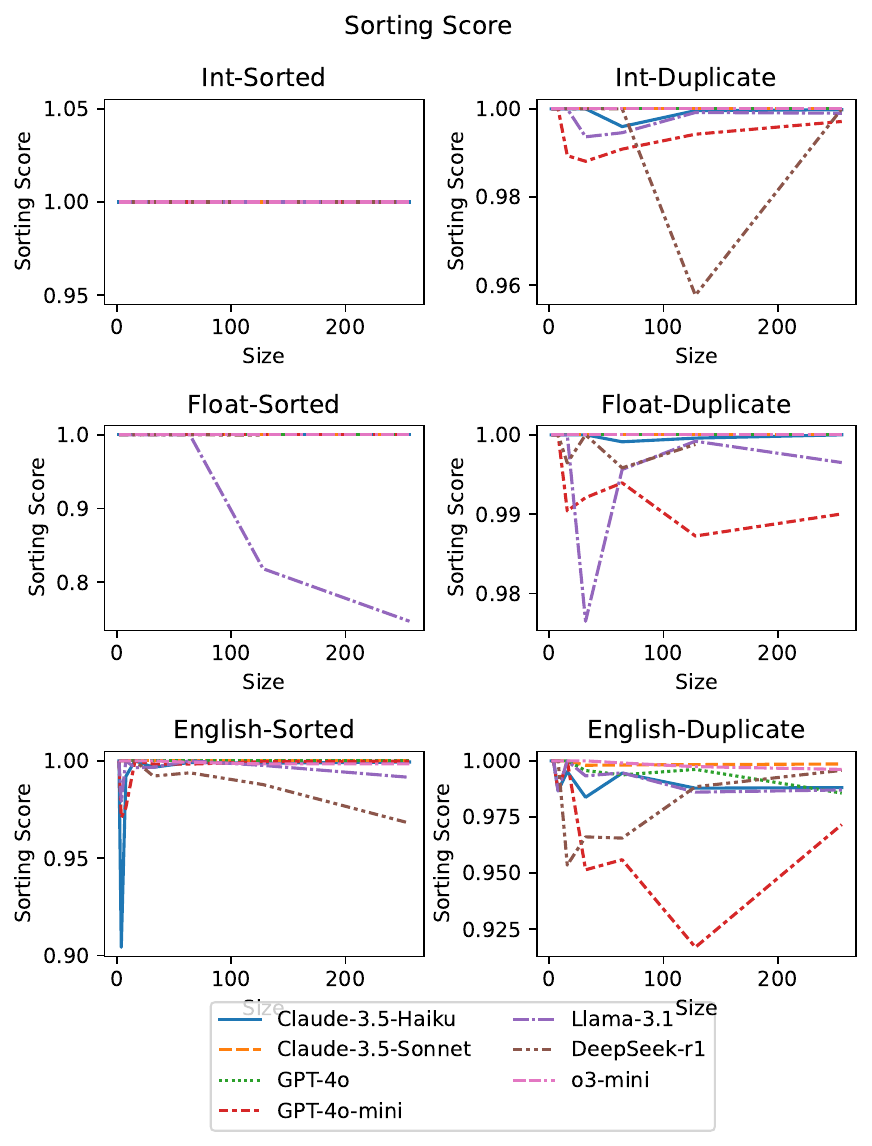}
\caption{$SortingScore$ for all debug tasks by list size}
\label{fig:results-debug-sort}
\end{figure}

\begin{figure}[h]
\centering
\includegraphics[width=0.6\linewidth]{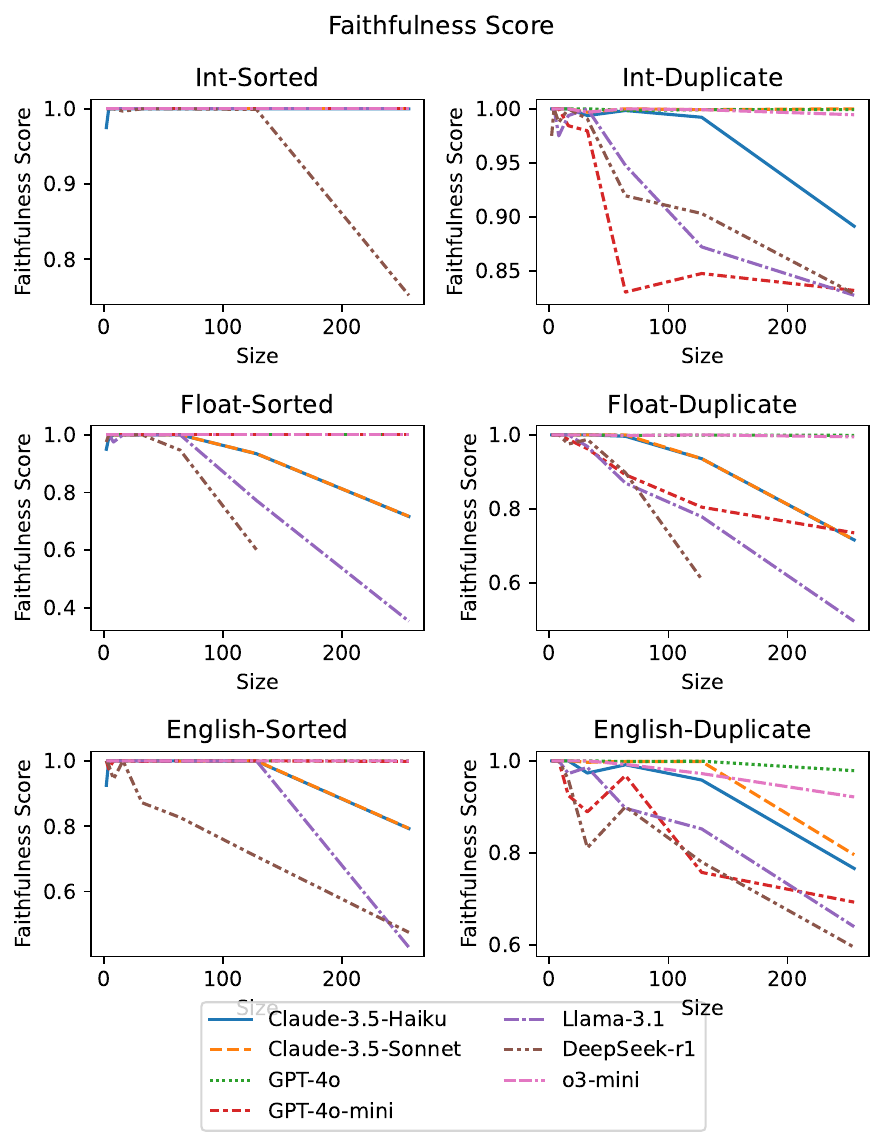}
\caption{$FaithfulnessScore$ for all debug tasks by list size}
\label{fig:results-debug-faith}
\end{figure}

Within this section, we provide a visual analysis of the trends for all tasks and scores. Figures~\ref{fig:results-basic-aggr}-\ref{fig:results-advanced-faith} show the results for the basic tasks, figures~\ref{fig:results-advanced-aggr}-\ref{fig:results-advanced-faith} for the advanced tasks, and figures~\ref{fig:results-debug-aggr}-\ref{fig:results-debug-faith} for the debug tasks. Each figure shows the trend line over the averages of the ten lists of the same length for each task.

\FloatBarrier

\subsection{Statistical analysis}
\label{sec:details-stats}

This section presents the results of the statistical analysis of the $SortBenchScore$ over all tasks for different list length. The Tables~\ref{tbl:stats-2}--\ref{tbl:stats-256} report the mean value (M), standard deviation (SD), the 95\% confidence interval (CI) of the mean value, Cohen's $d$ as effect size for the difference of means~\citep{cohen2013statistical} with respect to the best ranked model, and the probability that a model is smaller than the best ranked model ($p^{best}_{smaller}$) and the one reported in the line above ($p^{above}_{smaller}$), computed with a Bayesian signed rank test~\cite{benavoli2014bayesian} with a Region of Practical Equivalence (ROPE) as $\pm 0.1 \cdot d$ following \cite{kruschke2018bayesian}. The effect size is only reported in relation to the best-ranked model, but only if there is at least a 80\% probability, that a the mean performance is smaller than that of the best-ranked model. The computation of the confidence intervals uses Bonferroni-Dunn correction for the number of models to ensure that the family-wise error for each list length is 0.05. 

This reporting follows the guidelines by \cite{benavoli2017time}, and updated version of the popular guidelines by \cite{demvsar2006statistical}. 

\begin{table}[h]
\centering
\begin{tabular}{lcccccc}
\toprule
Model & M & SD & CI & $d$ & $p^{best}_{smaller}$ & $p^{above}_{smaller}$\\
\midrule
o3-mini & 1.000 & 0.000 & [1.000, 1.000] & - & - \\
Claude-3.5-Sonnet & 0.994 & 0.048 & [0.982, 1.006] & - & 0.000 & 0.000\\
GPT-4o & 0.992 & 0.055 & [0.978, 1.005] & - & 0.000 & 0.000\\
Llama-3.1 & 0.985 & 0.073 & [0.967, 1.004] & - & 0.000 & 0.000\\
GPT-4o-mini & 0.981 & 0.082 & [0.961, 1.002] & - & 0.000 & 0.000\\
Claude-3.5-Haiku & 0.896 & 0.187 & [0.850, 0.943] & - & 0.007 & 0.003 \\
DeepSeek-r1 & 0.867 & 0.232 & [0.809, 0.925] & - & 0.073 & 0.029 \\
\bottomrule
\end{tabular}
\caption{Statistics for lists of length 2 for all tasks}
\label{tbl:stats-2}
\end{table}

\begin{table}[h]
\centering
\begin{tabular}{lcccccc}
\toprule
Model & M & SD & CI & $d$ & $p^{best}_{smaller}$ & $p^{above}_{smaller}$\\
\midrule
o3-mini & 1.000 & 0.000 & [1.000, 1.000] & - & - & - \\
Claude-3.5-Sonnet & 0.997 & 0.017 & [0.993, 1.002] & - & 0.000 & 0.000 \\
GPT-4o & 0.996 & 0.026 & [0.990, 1.003] & - & 0.000 & 0.000 \\
GPT-4o-mini & 0.984 & 0.055 & [0.971, 0.998] & - & 0.000 & 0.000 \\
Llama-3.1 & 0.972 & 0.072 & [0.955, 0.990] & - & 0.000 & 0.000 \\
Claude-3.5-Haiku & 0.953 & 0.135 & [0.919, 0.986] & - & 0.000 & 0.000 \\
DeepSeek-r1 & 0.909 & 0.193 & [0.861, 0.957] & - & 0.009 & 0.004 \\
\bottomrule
\end{tabular}
\caption{Statistics for lists of length 4 for all tasks}
\label{tbl:stats-4}
\end{table}

\begin{table}[h]
\centering
\begin{tabular}{lcccccc}
\toprule
Model & M & SD & CI & $d$ & $p^{best}_{smaller}$ & $p^{above}_{smaller}$\\
\midrule
o3-mini & 1.000 & 0.000 & [1.000, 1.000] & - & - & - \\
Claude-3.5-Sonnet & 0.995 & 0.026 & [0.989, 1.002] & - & 0.000 & 0.000 \\
GPT-4o & 0.991 & 0.039 & [0.982, 1.001] & - & 0.000 & 0.000 \\
GPT-4o-mini & 0.977 & 0.047 & [0.965, 0.989] & - & 0.083 & 0.203 \\
Llama-3.1 & 0.964 & 0.082 & [0.944, 0.984] & - & 0.102 & 0.089 \\
Claude-3.5-Haiku & 0.932 & 0.162 & [0.891, 0.972] & - & 0.005 & 0.033 \\
DeepSeek-r1 & 0.906 & 0.217 & [0.852, 0.960] & - & 0.003 & 0.000 \\
\bottomrule
\end{tabular}
\caption{Statistics for lists of length 8 for all tasks}
\label{tbl:stats-8}
\end{table}

\begin{table}[h]
\centering
\begin{tabular}{lcccccc}
\toprule
Model & M & SD & CI & $d$ & $p^{best}_{smaller}$ & $p^{above}_{smaller}$\\
\midrule
o3-mini & 1.000 & 0.000 & [1.000, 1.000] & - & - & -  \\
Claude-3.5-Sonnet & 0.993 & 0.041 & [0.983, 1.003] & - & 0.000 & 0.000 \\
GPT-4o & 0.983 & 0.068 & [0.966, 1.000] & - & 0.000 & 0.000 \\
GPT-4o-mini & 0.945 & 0.127 & [0.913, 0.976] & 0.616 & 1.000 & 0.996 \\
Llama-3.1 & 0.940 & 0.138 & [0.906, 0.974] & 0.615 & 0.816 & 0.021 \\
Claude-3.5-Haiku & 0.932 & 0.158 & [0.893, 0.972] & - & 0.042 & 0.001 \\
DeepSeek-r1 & 0.827 & 0.292 & [0.754, 0.900] & - & 0.620 & 0.477 \\

\bottomrule
\end{tabular}
\caption{Statistics for lists of length 16 for all tasks}
\label{tbl:stats-16}
\end{table}

\begin{table}[h]
\centering
\begin{tabular}{lcccccc}
\toprule
Model & M & SD & CI & $d$ & $p^{best}_{smaller}$ & $p^{above}_{smaller}$\\
\midrule
o3-mini & 0.997 & 0.027 & [0.990, 1.003] & - & - & - \\
Claude-3.5-Sonnet & 0.987 & 0.044 & [0.976, 0.998] & - & 0.000 & 0.000 \\
GPT-4o & 0.978 & 0.074 & [0.959, 0.996] & - & 0.000 & 0.000 \\
Claude-3.5-Haiku & 0.937 & 0.139 & [0.902, 0.971] & - & 0.565 & 0.004 \\
GPT-4o-mini & 0.936 & 0.112 & [0.908, 0.964] & 0.750 & 1.000 & 0.965 \\
Llama-3.1 & 0.925 & 0.163 & [0.884, 0.966] & 0.613 & 1.000 & 0.018 \\
DeepSeek-r1 & 0.719 & 0.358 & [0.630, 0.808] & 1.097 & 1.000 & 1.000 \\
\bottomrule
\end{tabular}
\caption{Statistics for lists of length 32 for all tasks}
\label{tbl:stats-32}
\end{table}

\begin{table}[h]
\centering
\begin{tabular}{lcccccc}
\toprule
Model & M & SD & CI & $d$ & $p^{best}_{smaller}$ & $p^{above}_{smaller}$\\
\midrule
o3-mini & 0.996 & 0.025 & [0.990, 1.003] & - & - & - \\
Claude-3.5-Sonnet & 0.983 & 0.051 & [0.971, 0.996] & - & 0.000 & 0.000 \\
GPT-4o & 0.974 & 0.073 & [0.956, 0.992] & - & 0.007 & 0.002 \\
Claude-3.5-Haiku & 0.942 & 0.124 & [0.911, 0.973] & 0.610 & 0.985 & 0.003 \\
GPT-4o-mini & 0.907 & 0.143 & [0.871, 0.943] & 0.872 & 1.000 & 1.000 \\
Llama-3.1 & 0.897 & 0.152 & [0.860, 0.935] & 0.907 & 1.000 & 0.406 \\
DeepSeek-r1 & 0.742 & 0.318 & [0.662, 0.822] & 1.129 & 1.000 & 0.999 \\
\bottomrule
\end{tabular}
\caption{Statistics for lists of length 64 for all tasks}
\label{tbl:stats-64}
\end{table}

\begin{table}[h]
\centering
\begin{tabular}{lcccccc}
\toprule
Model & M & SD & CI & $d$ & $p^{best}_{smaller}$ & $p^{above}_{smaller}$\\
\midrule
o3-mini & 0.986 & 0.047 & [0.974, 0.998] & - & - & - \\
GPT-4o & 0.932 & 0.167 & [0.890, 0.974] & - & 0.001 & 0.001 \\
Claude-3.5-Sonnet & 0.859 & 0.157 & [0.819, 0.898] & 1.095 & 1.000 & 1.000 \\
GPT-4o-mini & 0.850 & 0.197 & [0.801, 0.899] & 0.950 & 1.000 & 0.578 \\
Llama-3.1 & 0.846 & 0.180 & [0.802, 0.891] & 1.057 & 1.000 & 0.809 \\
Claude-3.5-Haiku & 0.838 & 0.165 & [0.796, 0.879] & 1.217 & 1.000 & 0.885 \\
DeepSeek-r1 & 0.647 & 0.287 & [0.576, 0.719] & 1.646 & 1.000 & 1.000 \\
\bottomrule
\end{tabular}
\caption{Statistics for lists of length 128 for all tasks}
\label{tbl:stats-128}
\end{table}

\begin{table}[h]
\centering
\begin{tabular}{lcccccc}
\toprule
Model & M & SD & CI & $d$ & $p^{best}_{smaller}$ & $p^{above}_{smaller}$\\
\midrule
o3-mini & 0.963 & 0.072 & [0.945, 0.981] & - & - & - \\
GPT-4o & 0.873 & 0.220 & [0.819, 0.928] & 0.546 & 0.956 & 0.956 \\
GPT-4o-mini & 0.796 & 0.240 & [0.737, 0.856] & 0.940 & 1.000 & 1.000 \\
Claude-3.5-Sonnet & 0.716 & 0.159 & [0.676, 0.756] & 1.998 & 1.000 & 1.000 \\
Claude-3.5-Haiku & 0.702 & 0.167 & [0.661, 0.744] & 2.027 & 1.000 & 0.000 \\
Llama-3.1 & 0.528 & 0.362 & [0.437, 0.619] & 1.664 & 1.000 & 1.000 \\
DeepSeek-r1 & 0.345 & 0.366 & [0.253, 0.436] & 2.345 & 1.000 & 1.000 \\
\bottomrule
\end{tabular}
\caption{Statistics for lists of length 256 for all tasks}
\label{tbl:stats-256}
\end{table}

\end{document}